\pdfoutput=1
\documentclass[11pt]{article}

\usepackage{EACL2023}

\usepackage{times}
\usepackage{latexsym}

\usepackage[T1]{fontenc}
\DeclareUnicodeCharacter{25C7}{$\diamond$}
\DeclareUnicodeCharacter{27F6}{$\li$}
\DeclareUnicodeCharacter{25A1}{$\bx$}
\DeclareUnicodeCharacter{25B5}{$\diaintro$}
\DeclareUnicodeCharacter{03BB}{$\lambda$}
\DeclareUnicodeCharacter{25BE}{$\boxelim$}
\DeclareUnicodeCharacter{3008}{$\langle$}
\DeclareUnicodeCharacter{3009}{$\rangle$}

\usepackage[utf8]{inputenc}

\usepackage{microtype}

\usepackage{inconsolata}

\definecolor{color1}{HTML}{8b0000}
\definecolor{color2}{HTML}{006400}
\definecolor{termcolor}{HTML}{000000}

\usepackage{amssymb}
\usepackage{graphicx}
\usepackage{relsize}
\usepackage{ifthen}
\usepackage{tabularx}
\usepackage{booktabs}
\usepackage{amsmath}
\usepackage{proof}
\usepackage{stmaryrd}
\makeatletter
\def\new@fontshape{}
\makeatother
\usepackage{gb4e}
\noautomath
\usepackage{tikz}
\usetikzlibrary{calc,positioning,shapes.geometric,shapes.symbols,shapes.misc}
\usepackage{xfrac}
\usepackage[frozencache,cachedir=pygcache]{minted}

\DeclareUnicodeCharacter{25C7}{$\diamond$}
\DeclareUnicodeCharacter{27F6}{$\to$}
\DeclareUnicodeCharacter{25A1}{$\bx$}
\DeclareUnicodeCharacter{25B5}{$\diaintro$}
\DeclareUnicodeCharacter{03BB}{$\lambda$}
\DeclareUnicodeCharacter{25BE}{$\bxelim$}
\DeclareUnicodeCharacter{22A2}{$\vdash$}

\newcolumntype{C}{>{\centering\arraybackslash}X}
\newcommand{\doliscale}[2]{\raisebox{0.15\getfontdim{#1}}{\scalebox{0.9}{\ensuremath{#1#2}}}}
\newcommand{\liscale}[1]{\mathpalette\doliscale{#1}\relax}
\newcommand{\li}{\liscale{\multimap}}
\newcommand{\smallgtype}[1]{\ensuremath{\textsc{\normalfont\textsc{#1}}}}
\newcommand{\type}[1]{\smallgtype{#1}}
\makeatletter
\newcommand{\getfontdim}[1]{%
  \fontdimen22
  \ifx#1\displaystyle\textfont\else
  \ifx#1\textstyle\textfont\else
  \ifx#1\scriptstyle\scriptfont\else
  \scriptscriptfont\fi\fi\fi 2
}
\makeatother
\newcommand{\domodaldiamondscale}[2]{\raisebox{0.5\getfontdim{#1}}{\scalebox{0.9}{\ensuremath{#1#2}}}}
\newcommand{\domodalboxscale}[2]{\raisebox{0.15\getfontdim{#1}}{\scalebox{0.85}{\ensuremath{#1#2}}}}
\newcommand{\modaldiamondscale}[1]{\mathpalette\domodaldiamondscale{#1}\relax}
\newcommand{\modalboxscale}[1]{\mathpalette\domodalboxscale{#1}\relax}
\renewcommand{\diamond}{\modaldiamondscale{\diamondsuit}}
\newcommand{\bx}{\modalboxscale{\Box}}
\newcommand{\dep}[1]{\ensuremath{\mathsf{#1}}}
\newcommand{\ddia}[1]{%
    \diamond_{\scalebox{0.7}{\dep{\smaller{#1}}}}
}
\newcommand{\dbox}[1]{%
    \bx_{\scalebox{0.7}{\dep{\smaller{#1}}}}
}
\newcommand{\bracket}[1]{\langle{#1}\rangle}
\newcommand{\dbra}[2]{\bracket{#1}^{\scalebox{0.7}{\dep{#2}}}}
\newcommand{\abra}[2]{\dbra{#1}{\textcolor{color2}{#2}}}
\newcommand{\cbra}[2]{\dbra{#1}{\textcolor{color1}{#2}}}
\newcommand{\gtype}[2][l]{%
	\ifthenelse
		{\equal{#1}{s}}%
		{\type{#2}}%
		{\ensuremath{\vnorm{\type{#2}}}}%
}
\newcommand{\vnorm}[1]{\vphantom{\dbra{A}{X}}{#1}}
\newcommand{\subcat}[3]{\gtype[#1]{#2}\textsubscript{#3}}
\newcommand{\s}[1][]{\gtype[#1]{s}}
\newcommand{\smalls}{\s[s]}
\newcommand{\smain}[1][]{\subcat{#1}{\smalls}{main}}

\newcommand{\rulestyle}[1]{\ensuremath{\mathrm{#1}}}
\newcommand{\lex}{\rulestyle{lex}}
\newcommand{\ax}{\rulestyle{id}}
\newcommand{\w}[1]{\text{#1}}
\newcommand{\term}[1]{\ensuremath{\mathsf{#1}}}

\newcommand{\texttr}[2]{\textit{#1}~`#2'}

\makeatletter
\newcommand{\domodaltermscale}[2]{\raisebox{0.25\getfontdim{#1}}{\scalebox{0.85}{\ensuremath{#1#2}}}}
\makeatother
\newcommand{\modaltermscale}[1]{\mathpalette\domodaltermscale{#1}\relax}
\newcommand{\bxelim}[1][]{\modaltermscale{\blacktriangledown}_{\dep{#1}}}

\newcommand{\diaintro}[1][]{\modaltermscale{\vartriangle}_{\dep{#1}}}

\newcommand{\spindle}{spind\textsuperscript{2}$\lambda$e}

\title{SPINDLE: \\Spinning Raw Text into Lambda Terms with Graph Attention}

\author{
    Konstantinos Kogkalidis\textsuperscript{$\diamond$} \and
    Michael Moortgat\textsuperscript{$\diamond$} \and
    Richard Moot\textsuperscript{$\bx$} \\ 
    {$\diamond$} Institute for Language Sciences, Utrecht University \\ 
    {$\bx$} LIRMM, Universit\'{e} de Montpellier, CNRS \\
    \texttt{\{k.kogkalidis,m.j.moortgat\}@uu.nl, richard.moot@lirmm.fr}
}

\begin{document}
\maketitle
\begin{abstract}
This paper describes SPINDLE%
    \footnote{Stylized \spindle{} and standing for \textbf{s}pindle \textbf{p}arses \textbf{in}to \textbf{d}ependency-\textbf{d}ecorated $\mathbf{\lambda}$ \textbf{e}xpressions. 
    Source code and user instructions can be found at \url{https://github.com/konstantinosKokos/spindle}.} --
an open source Python module implementing an efficient and accurate parser for written Dutch that transforms raw text input to programs for meaning composition, expressed as $\lambda$ terms.
The parser integrates a number of breakthrough advances made in recent years.
Its output consists of hi-res derivations of a multimodal type-logical grammar, capturing two orthogonal axes of syntax, namely deep function-argument structures and dependency relations.
These are produced by three interdependent systems: a static type-checker asserting the well-formedness of grammatical analyses, a state-of-the-art, structurally-aware supertagger based on heterogeneous graph convolutions, and a massively parallel proof search component based on Sinkhorn iterations.
Packed in the software are also handy utilities and extras for proof visualization and inference, intended to facilitate end-user utilization.
\end{abstract}

\section{Introduction}
The transparency and formal well-behavedness of lambda calculi make them the ideal format for expressing compositional structures, a fact that has been duly emphasized by parsers and tools with a predominant focus on semantics.
Lambda calculi form a key ingredient of type-logical grammars, where they find use as the computational counterpart of a so-called \textit{grammar logic}, a substructural logic
of the intuitionistic linear variety that is designed to capture (aspects of) natural language syntax and semantics~\cite{moortgat1997categorial}.
For type-logical grammars, the Curry-Howard isomorphism guarantees a straightforward passage between logical rules, type constructors and term-forming operators; put simply,
Parse $\equiv$ Proof $\equiv$ Program, and Category $\equiv$ Proposition $\equiv$ Type.
The modus operandi is straightforward: a lexicon associates words with logical formulas, and the logic's rules of inference decide how formulas may interact with one another.

By extension, words may only combine in a strict, well-typed manner, forming larger phrases in the process.
Parsing becomes a process of logical deduction, at the end of which the result (a proof) gives rise to a recipe for meaning assembly (a program).
This program is turned into executable code as soon as one plugs in appropriate interpretations for the lexical constants (words) and for the term operations (composition instructions).
The set-up is general-purpose in that it readily accommodates different choices for these interpretations; valid targets can for instance be found in (truth-conditional) formal semantics, distributional-compositional models~\cite{DBLP:journals/jlm/SadrzadehM18a}, or tableau-based theorem provers~\cite{abzianidze-2017-langpro}.

In this work, we are interested in what happens prior to semantic execution; that is, we abstract away from lexical semantics and seek to reveal the compositional recipe underlying a natural language utterance.
To that end, we employ a type grammar aimed at capturing two different \textit{syntactic} axes, only rarely observed together in the wild:
function-argument structures and dependency relations.
To procure a derivation from an input phrase, we design and implement a system combining three distinct but communicating components.
Component number one is the implementation of the grammar's type system --- it comes packed with a number of useful facilities,
most important being a static type checker that verifies the syntactic well-formedness of the analyses construed.
Component number two is a supertagger responsible for assigning a type to each input word --- the tagger is formulated on the basis of a
hyper-efficient heterogeneous graph convolution kernel that boasts state-of-the-art accuracy among categorial grammar datasets.
The third and last component is a neural permutation module that exploits the linearity constraint of the target logic to simplify proof search as optimal transport learning~\cite{peyre2019computational} --- this reformulation allows for a massively parallel and easily optimizable implementation, unobstructed by the structure manipulation breaks common in conventional parsers.
The three components alternate roles through the processing pipeline, switching between phases of low level linear algebra routines and high level logical reasoning (GPU and CPU intensive, respectively).
Their integration yields a lightspeed-fast and highly accurate neurosymbolic parser, neatly packaged and made publicly available.

\section{System Decomposition}
\subsection{Type Grammar}
The system's theoretical backbone is its type logic -- a uniquely flavoured, semantics-first type-logical grammar that strays from the categorial norm in two major ways.
First, it focuses on deep syntactic structure (or tectogrammar, in~\citeauthor{curry1961some}'s terms) rather than surface form; its functional types are therefore oblivious to directional or positional constraints, abiding only to the linearity condition: every occurrence of an atomic proposition must be used once and exactly once.
Second, it dresses functional types up, so as to have them encode grammatical functions, making a three-way distinction between complements, heads and adjuncts.

A full exposition of the grammar is beyond the scope of this paper, but a superficial and simplified rundown should help shed light on what is to follow.
Its first aspect, \textbf{function-argument structures}, is modeled using linear logic's implication arrow, $\li$, which gives us access to resource-conscious versions of function application and variable abstraction~\cite{girard1987linear,abramsky1993computational}.
In their linguistic usecase, functional types of the form $\type{a}\li\type{b}$ denote predicates that \textit{consume} a single occurrence of some object of type $\type{a}$, the result being a composite phrase of type $\type{b}$.
Reasoning about gaps, ellipses and the like is accomplished with the aid of higher-order types, i.e. instances of the previous scheme where $\type{a}$ is itself a function --- these higher-order types launch a process of \textbf{hypothetical reasoning}, whereby we may temporarily assume the existence of a resource to produce a derivation locally, only to later withdraw the hypothesis, creating a new function in the process.
The second aspect, \textbf{dependency relations}, are modeled using a labeled assortment of residuated pairs of unary operators lent from temporal logic.
Atomic types without any dependency decorations are assigned to linguistically autonomous units and phrases, e.g. $\type{np}$ for a noun phrase.
Functional types denoting heads impose a diamond $\ddia{c}$ on the complements they select for, label $\dep{c}$ being the dependency slot the complement is to occupy, e.g. $\ddia{su}\gtype{np}\li\smain[s]$ for an intransitive verb looking for a subject-marked noun phrase to produce a matrix clause.
Dually, functional types denoting adjuncts are themselves decorated with a box $\dbox{a}$, label $\dep{a}$ now being the dependency role projected by the adjunct prior to application, e.g. $\dbox{mod}(\type{np}\li\type{np})$ for an adjective, promising to provide a function over noun phrases if one is to remove its box.
Introducing a diamond or eliminating a box leaves a \textit{structural} imprint that encloses complete phrases under brackets, and a \textit{computational} imprint that calls for a special treatment of the wrapped term -- both labeled by the grammatical function of the diamond (resp. box) that was introduced (resp. eliminated).
The key logical rules of the type grammar and their isomorphic term operations are presented in Figure~\ref{figure:inference_rules}.

\begin{figure}[t]
	\centering
	{\smaller
	\begin{tabularx}{0.49\textwidth}{@{}CC@{}}
	    $\infer[\ax]{\term{x}: \type{a} \vdash~\textcolor{black}{\term{x}}: \type{a}}{}$
	    &
	    $\infer[\lex]{\term{c}: \type{a} \vdash~\textcolor{black}{\term{c}}: \type{a}}{(\term{c}\mapsto \type{a}) \in \mathcal{L}}$\\[1em]
		$\infer[\li E]{\Gamma, \Delta \vdash \textcolor{black}{\term{s~t}}: \type{b}}{\Gamma \vdash \textcolor{black}{\term{s}}: \type{a} \li \type{b} & \Delta \vdash \textcolor{black}{\term{t}}: \type{a}}$
		& 
		$\infer[\li I]{\Gamma \vdash \textcolor{black}{\term{\lambda x.s}}: \type{a} \li \type{b}}{\Gamma, \term{x}: \type{a} \vdash \textcolor{black}{\term{s}}: \type{b}}$\\[1em]
		$\infer[\dbox{\delta} E]{\dbra{\Gamma}{\textcolor{color2}{\delta}} \vdash \textcolor{black}{\bxelim[\delta]{\term{s}}}: \type{a}}{
			\Gamma \vdash \textcolor{black}{\term{s}}: \dbox{\delta}\type{a}
		}$
		&
		$\infer[\ddia{\delta} I]{\dbra{\Gamma}{\textcolor{color1}{\delta}} \vdash \textcolor{black}{\diaintro[\delta]{\term{s}}}: \ddia{\delta} \type{a}}{
			\Gamma \vdash \textcolor{black}{\term{s}}:\type{a}
		}$
	\end{tabularx}
	}
	\caption{Logical rules of inference used by the type grammar (subset).
	The $\ax$ rule instantiates a fresh variable of arbitrary type $\type{a}$. 
    The $\lex$ rule provides declares a constant $\term{c}$ as being of type $\type{a}$, given type assignment $\term{c}\mapsto\type{a}$ pulled from the lexicon $\mathcal{L}$ (or, in the post-neural era, the supertagger).
    Introduction rules are complex types constructors, elimination rules are destructors.
    The $\li E$ rule says a term $\term{s}$ of type $\type{a}\li\type{b}$ derived from premises $\Gamma$ can apply to a term $\term{t}$ of type $\type{a}$ derived from premises $\Delta$, producing a complex term $\term{s~t}$ of type $\type{b}$ derived from the merger of $\Gamma$ and $\Delta$.
    The $\li I$ rule says that if the premises of some term $\term{s}$ of type $\type{b}$ include a variable $\term{x}$ of type $\type{a}$, we can abstract over the latter, producing a term $\term{\lambda x.s}$ of type $\type{a}\li\type{b}$.
    The $\dbox{\delta} E$ rule removes the box from a term $\term{s}$ of type $\dbox{\delta}\type{a}$, producing term $\bxelim[\delta]\term{s}$ of type $\type{a}$ and enclosing the premises under brackets $\dbra{\_}{\delta}$.
    Dually, the $\ddia{\delta}I$ rule puts a term $\term{s}$ of type $\type{a}$ under the scope of a diamond, producing term $\diaintro[\delta]{\term{s}}$ of type $\ddia{\delta}{\type{a}}$ and again enclosing the premises under brackets $\dbra{\_}{\delta}$.}
	\label{figure:inference_rules}
\end{figure}

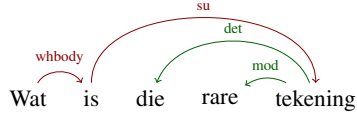
\begin{figure*}[t]
    \begin{tikzpicture}[t/.style={text depth=.25ex, rectangle, outer sep=0pt}]
        \node[t] (proof) at (0,0) {
            \resizebox{0.99\textwidth}{!}{
                \infer[\li E]
                    {\term{c_0}, \cbra{\term{c_1}, \cbra{\abra{\term{c_2}}{det},\abra{\term{c_3}}{mod}, \term{c_4}}{su}}{whbody} \vdash \textcolor{termcolor}{\term{c_0}~\diaintro[whbody](\lambda \term{x}.\term{c_1}~\term{x}~\diaintro[su](\bxelim[det](\term{c_2})~(\bxelim[mod](\term{c_3})~\term{c_4}))}:
 \gtype{whq}}
                    {
                        \infer[\lex]{\term{c_0}\vdash \ddia{whbody}(\ddia{predc}\gtype{pron}\li\gtype{svi})\li\gtype{whq}}{\w{Wat}}
                        &
                        \hspace{-105pt}
                        \infer[\ddia{whbody}I]
                        {\cbra{\term{c_1}, \abra{\abra{\term{c_2}}{det},\abra{\term{c_3}}{mod}, \term{c_4}}{su}}{whbody} \vdash \ddia{whbody}({\ddia{predc}\gtype{pron}\li\gtype{svi}})}
                        {
                            \infer[\li I]
                            {\term{c_1}, \cbra{\abra{\term{c_2}}{det},\abra{\term{c_3}}{mod}, \term{c_4}}{su} \vdash \ddia{predc}\gtype{pron}\li\gtype{svi}}
                            {
                                \infer[\li E]
                                {\term{c_1}, \term{x}, \cbra{\abra{\term{c_2}}{det},\abra{\term{c_3}}{mod}, \term{c_4}}{su} \vdash \gtype{svi}}
                                {
                                    \infer[\li E]
                                    {\term{c_1}, \term{x} \vdash \ddia{su}\gtype{np}\li\gtype{svi}}
                                    {
                                        \infer[\lex]
                                        {\term{c_1}\vdash \ddia{predc}\gtype{pron}\li\ddia{su}\gtype{np}\li\gtype{svi}}
                                        {\w{is}}
                                        &
                                        \infer[\ax]
                                        {\term{x}\vdash \ddia{predc}\gtype{pron}}
                                        {}
                                    }
                                    &
                                    \hspace{-10pt}
                                    \infer[\ddia{su} I]
                                    {\cbra{\abra{\term{c_2}}{det},\abra{\term{c_3}}{mod}, \term{c_4}}{su} \vdash \ddia{su}\gtype{np}}
                                    {
                                        \infer[\li E]
                                        {\abra{\term{c_2}}{det},\abra{\term{c_3}}{mod}, \term{c_4} \vdash \gtype{np}}
                                        {
                                            \infer[\dbox{det} E]
                                            {\abra{\term{c_2}}{det} \vdash \gtype{n}\li\gtype{np}}
                                            {
                                                \infer[\lex]
                                                {\term{c_2}\vdash \dbox{det}(\gtype{n}\li\gtype{np})}
                                                {\w{die}}
                                            }
                                            &
                                            \infer[\li E]
                                            {\abra{\term{c_3}}{mod}, \term{c_4} \vdash \gtype{n}}
                                            {
                                                \infer[\dbox{mod} E]
                                                {\abra{\term{c_3}}{mod} \vdash \gtype{n}\li\gtype{n}}
                                                {
                                                    \infer[\lex]
                                                    {\term{c_3}\vdash \dbox{mod}(\gtype{n}\li\gtype{n})}
                                                    {\w{rare}}
                                                }
                                                &
                                                \infer[\lex]
                                                {\term{c_4}\vdash \gtype{n}}
                                                {\w{tekening}}
                                            }
                                        }
                                    }
                                }
                            }
                        }
                    }
                }
    	    };
    	    \node[t] (dep) at ($(proof.south) + (0, -1.53)$) {
    	        \resizebox{140pt}{!}{
                \begin{tikzpicture}
                \node[t] (wat) 			at (0, 0) {\w{Wat}};
                \node[t] (is)		    [right=10pt of wat] {\w{is}};
                \node[t] (dat) 			[right=10pt of is] {\w{die}};
                \node[t] (rare) 	    [right=10pt of dat] {\w{rare}};
                \node[t] (tekening)		[right=10pt of rare] {\w{tekening}};
                \draw[->,color=color1] (wat)         [bend left=60] edge node [above] {\smaller[2]{whbody}} (is);
                \draw[->,color=color1] (is)          [bend left=90] edge node [above] {\smaller[2]{su}} (tekening);
                \draw[->,color=color2] (tekening)      [bend right=30] edge node [above] {\smaller[2]{mod}} (rare);
                \draw[->,color=color2] (tekening)      [bend right=70] edge node [above] {\smaller[2]{det}} (dat);
                \end{tikzpicture}
                }
            };
	    \end{tikzpicture}
    \caption{Natural deduction proof for the sentence \texttr{Wat is die rare tekening?}{What is that strange drawing?}. For space economy, compositional $\lambda$ term is only explicitly written in the endsequent (bottom of the proof).
    From the antecedent structure of the endsequent, we may also recover a dependency tree. Color coding serves to informally differentiate between complement (red) vs. adjunct (green) structural brackets/dependency arcs.}
    \label{figure:nd_derivation}
\end{figure*}

\subsubsection{Proof Representation}
\begin{figure*}
    \resizebox{1\textwidth}{!}{
		\begin{tikzpicture}
		    [t/.style={text height=1.5ex, text depth=.25ex, rectangle, outer sep=0pt}, node distance=10pt,
		    r/.style={text=color1},
		    g/.style={text=color2},
		    tree/.style={very thick},
		    link/.style={dashed}]
		\node[t] (wat)      at (0, 0) {\w{Wat}};
		\node[t,g] (wat_fn_1) at (0, 1) {$\li\ddia{whbody}$};
        \node[t,r] (wat_fn_2) at (-1.5, 2.5) {$\li\ddia{predc}$};
        \node[t, g] (wat_pron) at (-2.5, 4) {$\type{pron}^0$};
        \node[t, r] (wat_svi) at (-0.5, 4) {$\subcat{s}{s}{vi}^1$};
        \node[t, g] (wat_whq) at (1.5, 2.5) {$\type{whq}^2$};

		\node[t] (is) at (4, 0)     {\w{is}};
		\node[t,g] (is_fn_1) at (4, 1)  {$\li\ddia{predc}$};
		\node[t,g] (is_fn_2) at (5, 2.5) {$\li\ddia{su}$};
		\node[t,r] (is_pron) at (3, 2.5) {$\type{pron}^3$};
		\node[t,r] (is_np) at (4.25, 4) {$\type{np}^4$};
		\node[t,g] (is_svi) at (5.75, 4) {$\subcat{s}{s}{vi}^5$};
		
		\node[t] (dat)   at (8, 0)   {\w{die}};
		\node[t,g] (dat_fn) at (8, 1)  {$\dbox{det}\li$};
		\node[t,r] (dat_n) at (7, 2.5) {$\type{n}^6$};
        \node[t,g] (dat_np) at (9, 2.5) {$\type{np}^7$};
        
        \node[t] (rare)   at (11, 0)   {\w{rare}};
		\node[t,g] (rare_fn) at (11, 1)  {$\dbox{mod}\li$};
		\node[t,r] (rare_n_1) at (10, 2.5) {$\type{n}^8$};
        \node[t,g] (rare_n_2) at (12, 2.5) {$\type{n}^9$};	
        
        \node[t] (tekening) at (14, 0) {\w{tekening}};
        \node[t,g] (tekening_n) at (14, 1) {$\type{n}^{10}$};

        \draw[tree] (wat_fn_1) -- (wat_fn_2) -- (wat_pron);
        \draw[tree] (wat_fn_2) -- (wat_svi);
        \draw[tree] (wat_fn_1) -- (wat_whq);
        
        \draw[tree] (is_fn_1) -- (is_pron);
        \draw[tree] (is_fn_1) -- (is_fn_2) -- (is_np);
        \draw[tree] (is_fn_2) -- (is_svi);
        
        \draw[tree] (dat_fn) -- (dat_n);
        \draw[tree] (dat_fn) -- (dat_np);
        
        \draw[tree] (rare_fn) -- (rare_n_1);
        \draw[tree] (rare_fn) -- (rare_n_2);

        \draw[->, dashed] (tekening_n) -- ($(tekening_n) + (0, 2.5)$) -| (rare_n_1);
        \draw[->, dashed] (rare_n_2) -- ($(rare_n_2) + (0, 1.5)$) -| (dat_n);
        \draw[->, dashed] (dat_np) -- ($(dat_np) + (0, 2.5)$) -| (is_np);
        \draw[->, dashed] (is_svi) -- ($(is_svi) + (0, 1.5)$) -| (wat_svi);
        \draw[->, dashed] (wat_pron) -- ($(wat_pron) + (0, 0.75)$) -| (is_pron);
        \draw[->, dashed] (wat_whq) -- ++ (0, 4);
%         \draw[->] (de_np) -- ($(de_np) + (0, 2)$) -| (waarover_np_1);
%         \draw[->] (waarover_np_2) -- ($(waarover_np_2) + (0, 1)$) -| (zijn_np);
%         \draw[->] (dit_prn) -- ($(dit_prn) + (0, 4.5)$) -| (zijn_prn);
%         \draw[->] (zijn_smain) -- ($(zijn_smain) + (0, 2)$);
%         \draw[->] (wij_prn) -- ($(wij_prn) + (0, 4)$) -| (staan_prn);
%         \draw[->] (waarover_adv) -- ($(waarover_adv) + (0, 1.5)$) -| (staan_adv);
%         \draw[->] (staan_ssub) -- ($(staan_ssub) + (0, 2)$) -| (waarover_ssub);
	    \end{tikzpicture}
    }
    \caption{Proof net equivalent of the proof of Figure~\ref{figure:nd_derivation}, with unary diamonds (resp. boxes) fused with the implication dominating (resp. dominated by) them for depth compression.
    Atomic propositions are indexed by enumeration for identification purposes.
    Color coding here serves to differentiate between resources we have (green) and resources we need (red) -- the rule is start green from the bottom, change (resp. keep) color for the left (resp. right) daughter of an implication.
    Bold edges denote the tree structure underlying type assignments. 
    Dashed edges denote the correct matching between resources of opposite polarity.}
    \label{figure:proofnet}
\end{figure*}
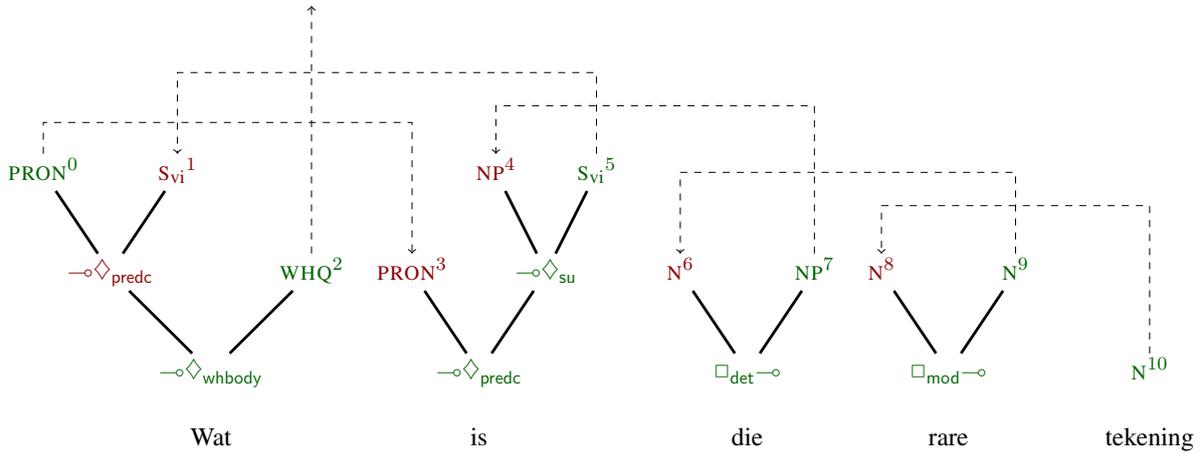

Proofs in the type logic are traditionally served in the tree-like \textbf{natural deduction} format.
Proofs in natural deduction benefit from an easy translation to (i) $\lambda$ expressions, by following the rules of Figure~\ref{figure:inference_rules}, and (ii) dependency trees, by simply casting structural brackets to dependency arcs, going from the head of each phrase to (the heads of) its dependents.
Figure~\ref{figure:nd_derivation} presents a visual example.
An alternative representation is in the far less verbose format of a \textbf{proof net}, a geometric construction that abstracts away from the bureaucratic book-keeping of hypothetical reasoning and tree-structured rule ordering.
Figure~\ref{figure:proofnet} presents the proof net equivalent of the running example.
Proof nets are easier to reason about in a neural setup by allowing us to treat parsing as the vastly simplified problem of matching each occurrence of an atomic proposition in \textit{negative} position, i.e. a prerequisite of a conditional implication, with an occurrence in \textit{positive} position, i.e. a (conditionally) proven statement.
The parallel nature of proof nets allows the matching to occur simultaneously across the entire proof; that is, all decisions are done in a single instant, without the bottleneck of having to wait for conditionals to be satisfied in a bottom-up fashion.
On the other hand, proof nets are slightly underspecified compared to natural deduction proofs, being explicit only with respect to function-argument structures -- translating from one format to another requires establishing some conventions on what constitutes a canonical proof.

\subsubsection{Implementation}
The syntax of the type system is implemented as a tiny DSL written in Python.\footnote{Source code can be found at \url{https://github.com/konstantinosKokos/aethel}.}
It is used as the representation format of \AE thel~\cite{kogkalidis-etal-2020-aethel}, a dataset of some 70\,000 analyses of written Dutch, which also constitutes the system's training data.
The implementation was originally designed to assert the type-safety of the dataset, to facilitate the conversion between natural deduction trees, $\lambda$ terms and proof nets, and to ease third-party corpus analysis by providing niceties such as search and pretty printing utilities, cross compilation to \LaTeX{} for visualization purposes, interfaces for proof transformations, etc.
All these functionalities are imported unchanged.
The conversion routines allow us to conduct neural proof search in the favorable regime of proof nets, and convert the result to natural deduction format only at the very end, just for the sake of presentation and/or sanity testing.
Importantly, the type-checker is repurposed as a tool for verifying the correctness of analyses constructed -- an analysis that does not amount to a valid proof will fail to pass the checker, throwing a type error and alerting us to the fact.
In other words, we can blindly trust anything the parser gives us as correct, at least in the sense of (proof-theoretic) syntactic validity.

\subsection{Supertagging Module}
Lexical type ambiguity and lexical type sparsity are common and pervasive problems for any categorial grammar.
The de facto approach rests on a supertagger, a neural module replacing the fixed lexicon, traditionally formulated as a sequence classifier and trained to produce the most plausible type assignments for each word in the context of an input sentence.
Here, these problems are exacerbated by the highly elaborated type system.
Some 80\% of \AE thel's approx. 6\,000 types are rare (i.e. have less than 10 occurrences in the corpus), and some 10\% of the total sentences contain at least 1 such rare type.
This necessitates a more ambitious treatment than the standard "set-and-forget" approach of completely discarding rare type assignments as inconsequential.
The solution comes in the form of a \textbf{constructive supertagger}, an auto-regressive neural decoder that is trained to construct types on the fly according to their algebraic decomposition, rather than treat them as singular, opaque blocks~\cite{kogkalidis-etal-2019-constructive}.
This configuration enables the construction of valid types regardless of whether they have been seen before or not, extending coverage beyond the training data.
The supertagger employed here follows a geometrically informed, task-specific decoding order, whereby types are represented as the structural unfolding of binary trees.
Following~\citet{prange-etal-2021-supertagging}, trees are decoded in parallel across the entire batch of input sequences, establishing an upper temporal bound on decoding that scales with the maximal tree depth -- in practice, a constant.
To circumvent the locality of a standard tree decoder, the target output being not a batch of trees but a batch of \textbf{sequences of trees} (see Figure~\ref{figure:proofnet}), the supertagger is formulated as a a graph neural network, utilizing message-passing connections to transfer feedback from tree nodes to their lexical roots and from lexical roots to their neighbours, ensuring that decisions made at each decoding step are influenced by prior decisions across the entire output~\cite{kogkalidis2022geometryaware}.
As a result, it strikes the perfect balance between the speed and memory efficiency of a tree-shaped architecture, allowing for more training iterations and faster inference, and the stronger autoregressive properties of a seq2seq model, improving performance.
Further, being inherently constrained to trees, its output is structurally correct-by-construction -- under no circumstance can any of the types produced be ill-formed.
Used in isolation, the architecture currently sits at the top of the accuracy leaderboard for categorial grammar supertagging across different formalisms and languages -- the performance is marginally inferior in the multi-task training setup adopted here.

\subsection{Permutation Module}
Conducting search over proof nets is typically ill-advised.
The problem traditionally involves examining all possible bijections between positive and negative atomic propositions.
The number of such bijections scales factorial to the number of atomic propositions, quickly becoming prohibitive.
To navigate this combinatorially explosive landscape, \textbf{neural proof nets} relax proof search into a continuous, differentiable problem, where finding the correct bijection is translated to a transportation problem~\cite{kogkalidis-etal-2020-neural} learned by yet another graph neural network.
Concretely, the representations of all occurrences of atomic propositions are extracted from the decoder and binned according to their sentential index, sign and polarity (e.g. a single bin would be all occurrences of a positive $\type{np}$ in sentence \#13 of the input batch).
Each bin is contrasted with its inverse polarity counterpart using some similarity metric (here a weighted inner product).
The result is a collection of square matrices, each matrix containing attention weights (or similarity scores) in the cartesian product of positive and negative items of the same sign and sentence.
These matrices are grouped by their cardinality, and the Sinkhorn operator~\cite{sinkhorn1967concerning} is used to push them towards binarity and bistochacity, yielding approximations of permutation matrices representing the goal bijections~\cite{mena2018learning}.

To make things concrete using the running example of Figure~\ref{figure:proofnet}, each of the atomic types $\subcat{s}{s}{vi}$, $\type{np}$ and $\type{pron}$ has a single negative and a single positive occurrence, therefore their bijections are trivial (a testament to supertagging being almost parsing).
The single positive occurrence of $\type{whq}$ stands for the goal type of the phrase, and remains unmatched.
Only $\type{n}$ requires a decision, having two possible bijections.
The correct candidate is encoded by the permutation table below, where rows enumerate positive and columns negative items:
\begin{center}
\begin{tikzpicture}[scale=.6]
    \fill[gray!30] (0,1) rectangle +(1,1);
    \node [above] at (0.5, 1) {\checkmark};
    \fill[gray!30] (1,0) rectangle +(1,1);
    \node [above] at (1.5, 0) {\checkmark};
    \node [above] at (0.5,2) {6};
    \node [above] at (1.5,2) {8};
    \node [left] at (0,1.5) {9};
    \node [left] at (0,0.5) {10};
    \node at (-1.75,1) {${\Pi}_\type{n}\!:=$};
    \draw (0,0) grid (2,2);
\end{tikzpicture}\hphantom{1500000}
\end{center}
This reformulation entails a tremendous speedup: the painstakingly slow problem of symbolic proof search is cast into simple, well-optimized and batchable matrix operations.
The current parser builds on the insight that the permutation module is invariant to the source of atomic symbol representations, and in fact greatly benefits from the faster and more accurate task-adapted supertagger.

\subsection{Integration}
Neurosymbolic integration yields an end-to-end pipeline that consists of the following phases.
First, the user inputs a list of sentences to be parsed.
Contextualized token representations are obtained from a fine-tuned BERT\textsubscript{BASE} model, which are then aggregated according to the input's word boundaries.
The resulting word representations act as initial seeds for decoding to begin on an empty canvas.
During decoding, types are progressively constructed, while seeds are updated by exchanging messages with one another on the basis of their sequential proximity.
After a small number of decoding steps, the process terminates, yielding a sequence of type assignments for each input sentence.
A rudimentary invariance check is then performed, controlling whether each sequence counts an equal number of atomic propositions of each polarity.
Sentences failing the invariance check are not eligible for proof search, and their analysis stops early.
Passing sentences are symbolically processed to obtain a collection of sparse indexing tensors, used to gather the decoder's representations into the bins described earlier.
Bins are batched and contrasted, and a small number of Sinkhorn iterations is employed as a 2-dimensional softmax analogue.
The soft Sinkhorn distances are discretized using the Hungarian algorithm in order to enforce bijectivity~\cite{jonker1987shortest}.
Bijections are reassociated with their origin symbols and sentences, using the reverse of the previous indexing operation.
Control is then passed to the symbolic component, which attempts to traverse the candidate proof nets, verifying the correctness criteria of acyclicity and connectedness in the process~\cite{danos1989structure}.
The traversal coincides with a translation to a natural deduction format, the construction of which corresponds to static type checking of the output~\cite{lamarche2008proof}.
Assuming no type mismatches are caught, the output is a proof proper, which by Curry-Howard isomorphism is rewritten as a dependency-decorated $\lambda$ term.
The user is finally presented with an analysis for each input sentence -- ideally, a $\lambda$ term, but occasionally a rejected intermediate result together with an error description.

\section{Evaluation}
\subsection{Performance}
The system has been evaluated on the test set of \AE thel.
Without any pre-filtering or post-processing training wheels (i.e. no constraints on sentence length, type rarity/depth or cardinality of bijections), the parser produces a proof that satisfies strict syntactic equality with the ground truth in 3\,191 of the 5\,770 test set samples.
This amounts to a significant 55.30\% of the test sentences being analyzed without a single error with respect to type assignments, phrasal chunking, function-argument structures and dependency annotations produced%
    \footnote{This is comparable to the state-of-the-art for the similar (in fact simpler) problem of CCG parsing; see \citet{DBLP:journals/corr/abs-2109-10044} for an up-to-date overview.}
In total, 4\,901 sentences are assigned a passing analysis, which sets the coverage to a more modest 84.94\%.
The discrepancy between the high accuracy and low coverage is due to the rigidness of the type system: only 5\,010 of the sentences satisfy the invariance check, being thus amenable to any proof.
This signals that the performance bottleneck lies on the supertagger rather than the permutation module; a parse is assigned to 97.82\% of parsable sentences, and it's also the perfect parse 75.30\% of the time.
These findings are summarized in Table~\ref{table:key_results}.

\begin{table}[h]
	\centering
	\begin{tabular}{@{}c@{\qquad}c@{}}
		\textbf{parsability}	&
		\textbf{coverage} \\
		\smaller(some proof obtainable) &
		\smaller(some proof obtained) \\
		\toprule
		86.83 &
		84.94 \\
		\addlinespace
		\textbf{types correct} &
		\textbf{accuracy}\\
		\smaller(correct proof obtainable) & 
		\smaller(correct proof obtained)\\
		\toprule
		56.88 & 
		{55.30}\\
	\end{tabular}
	\caption{Sentential-level evaluation of the parser.}
	\label{table:key_results}	
\end{table}

To obtain a more refined perspective on performance, we employ an adaptation of the parsing community's favorite $F_1$-score.
Concretely, we gather all samples for which a proof was produced, and decompose both prediction and ground truth into their respective sets of subproofs.
We measure $tp$ as the two sets' intersection, $fp$ as the difference between predicted and correct subproofs and $fn$ as the difference between correct and predicted subproofs, from which we may obtain precision as $p = \sfrac{tp}{(tp+fp)}$, recall as $r = \sfrac{tp}{(tp+fn)}$ and their harmonic mean as $F_1 = \sfrac{2 p r}{(p + r)}$.
On top of the vanilla versions of these metrics, we can also examine relaxations by incorporating a combination of two modulo factors.
Relaxation one targets the functional core of the logic, applying a forgetful transformation that strips proofs of their modalities in order to examine typed function-argument structures in isolation.
Relaxation two targets the modal enhancement of the logic, collapsing the set of atomic types into a single point (thus treating all functional types of the same \textit{shape} as equal) in order to examine dependency structures in isolation.
Relaxing on both axes at once is essentially casting proofs into the untyped $\lambda$ calculus, where all we care about are the type- and dependency- agnostic function-argument structures -- this is the metric most comparable to external theories.%
	\footnote{Proofs are in $\beta$ and $\eta$ normal, so no free points from abstractions. Variables are only equal if they match in both name and type, so no free points from variable instantiations either.}
Note that relaxations are performed only \textit{after} inference -- the point being that a strict proof must have been produced for its relaxations to be considered (i.e. lax accuracy is still bottlenecked by strict coverage).
The results are averaged over covered samples%
	\footnote{Averaging over the full test set would artificially inflate $p$ and deflate $r$ scores, since no partial proofs are returned from failing samples.}
and presented in Table~\ref{table:relaxations}.

\begin{table}[h]
	\centering
	\smaller
	\begin{tabular}{@{}l@{\qquad}ccc@{}}
		& \multicolumn{3}{c}{\textbf{local metrics}}\\
		\textbf{modulo}			& $p$ & $r$ & $F_1$\\
		\toprule
		--						& 89.52 & 89.68 & 89.39\\
		~modalities				& 90.93 & 91.13 & 90.85\\
		~functional types 		& 91.09 & 91.26 & 90.97\\
        ~both                   & 92.31 & 92.52 & 92.24
	\end{tabular}
	\caption{Decomposition metrics and relaxations.}
	\label{table:relaxations}	
\end{table}

\begin{figure*}
\smaller[2]
\begin{minted}[python3=true,texcomments]{pycon}
>>> from inference import InferenceWrapper as IW
>>> from aethel.utils.tex import compile_tex, sample_to_tex
>>> parser = IW(weight_path='./data/model_weights.pt')
>>> analysis = parser.analyze(['Wat is die rare tekening?'])[0]
>>> analysis
Analysis(
    lexical_phrases=(
        LexicalPhrase(string=Wat, type=(◇whbody(◇predc(VNW)⟶SV1))⟶WHQ, len=1),
        LexicalPhrase(string=is, type=◇predc(VNW)⟶◇su(NP)⟶SV1, len=1),
        LexicalPhrase(string=die, type=□det(N⟶NP), len=1),
        LexicalPhrase(string=rare, type=□mod(N⟶N), len=1),
        LexicalPhrase(string=tekening, type=N, len=1),
        LexicalPhrase(string=?, type=PUNCT, len=1)),
    proof=c0, 〈c1, 〈〈c2〉det, 〈c3〉mod, c4〉su〉whbody ⊢ c0 ▵whbody((λx0.c1 x0 ▵su(▾det(c2) (▾mod(c3) c4)))) : WHQ)
>>> tex_proof = compile_tex(sample_to_tex(analysis))  # see Figure \ref{figure:nd_derivation} for output :)
\end{minted}
    \caption{User interaction example in python console.}
    \label{figure:ui}
\end{figure*}

\subsection{Efficiency}
Regarding efficiency, the architecture contains a non-negligible total of 117M parameters, 94\% of which are inherited from the underlying BERT model.
The memory footprint of the network's forward pass does not exceed 3.5GB on the test set with a batch size of 64, making it reasonably lightweight for use at home.
Using a middle range laptop GPU, the network takes about 15 seconds to tag the full test set (i.e. 370 sents/sec or 6\,000 tokens/sec), and 80 seconds to tag and parse it (i.e. 70 sents/sec), including tokenization and post-processing.
Note, however, that sentence-level batching is not yet implemented for inference mode proof search, i.e. sentences are tagged in parallel but proven sequentially.
Cross-sentential padding and batching of Sinkhorn inputs is in the works -- benchmarking shows that the asymptotic behavior of a forward pass over batches of 64 matrices only starts becoming apparent when they exceed the order of $2^{7}$, being locked at an insubstantial 1ms prior to that.%
    \footnote{To comprehend how extremely unrealistic $2^7$ is, consider that this would amount to finding the correct bijection out of ${2^7}! = 3.9\times 10^{215}$ possibilities across 64 pairs of sets in parallel.}

\section{User Interface}
The user interface is bare-bones, but simple and easy to use.
The parser is currently packaged as a code repository, which, once downloaded, can be used as a python module.
A front-end class wraps around the scary inner workings of the parser and provides easy access to an inference routine.
Structure checking is handled internally and error handling is graceful: the user is guaranteed an output even in the case of a partial failure.
The output implements the same protocols as samples of the \AE thel corpus, and is thus compatible with all of the latter's bells and whistles.
Proofs can be pretty-printed, interactively processed and transformed (e.g. for semantic applications), or visualized using \LaTeX{} as a middlewoman.
For the more ambitious, training and evaluation utilities are also available.

\section{Conclusion \& Future Work}
Thus concludes the demonstration tour of \spindle{}: a unique neurosymbolic parser that can accurately and efficiently convert raw text into $\lambda$ expressions.
Unlike cheaper alternatives, these $\lambda$ expressions are not structureless ad-hoc imitations produced from arbitrary decoding, but executable, type-safe and 100\% \textbf{guaranteed correct} programs.
The software focuses on Dutch, but the universality of the intuitionistic linear core allows easy cross-lingual adaptation that essentially boils down to retraining with a new type lexicon; a French implementation is currently in the works~\cite{deepgrail23}.

As to what the future holds, the intention is to keep the module synchronized and up-to-date with \AE thel: any upcoming major release of the latter will be reflected in an update of the former (be it soft patching or retraining).
Compatibility aside, planned features include deploying the module as a web service, compiling it as a stand-alone package and documenting the annotations (so as to be more inclusive towards the type-uninitiated).
Any performance, stability or efficiency improvements stemming from related research or moments of engineering inspiration are also likely to find their way to the user-facing front.
Contributions and feedback are always welcome.
\goodbreak

\section*{Limitations}
The implementation described capitalizes on a disentanglement between neural and symbolic operations to improve efficiency.
But doing so comes at the heavy price of a unidirectional data flow that lacks feedback.
The symbolic component has the singular role of testing and verifying the neural output, but emits back no messages of its own.
Failures may be caught, but they are nonetheless irrecoverable -- a partial output that fails some structural constraint signifies an abrupt and non-negotiable end to the processing pipeline, significantly reducing coverage.
A better operationalization would be to use the symbolic core to continuously ask for neural output as long as the structural constraints are not met (or the user is not satisfied with the parse provided).
However, this would only be feasible if the neural components were to be extended with some notion of backtracking.
In that sense, the parallel nature of both the supertagger and the parser becomes now a double-edged sword, hindering the potential applicability of standard heuristic algorithms like beam search.

More generally, the software carries the standard risks of any NLP architecture reliant on machine learning, namely linguistic biases inherited from the unsupervised pretraining of the incorporated language model and annotation biases derived from the supervised training over human-labeled data.

\section*{Acknowledgements}
This software described was developed with funds from the Dutch Research Council (NWO, grant nr. 360-89-070).

\nocite{fey2019fast}

\bibliography{anthology, custom}
\bibliographystyle{acl_natbib}

\end{document}